\documentclass[runningheads]{llncs}
\usepackage{graphicx}

\usepackage{tikz}
\usepackage{comment}
\usepackage{amsmath,amssymb}
\usepackage{color}
\usepackage{bbm}

\usepackage[width=122mm,left=12mm,paperwidth=146mm,height=193mm,top=12mm,paperheight=217mm]{geometry}

\usepackage{color}
\usepackage{multirow}

\usepackage{bm}
\usepackage{siunitx}
\usepackage{url}
\usepackage{booktabs}
\usepackage{tabularx}
\usepackage{adjustbox}
\usepackage{subfigure}
\definecolor{turquoise}{cmyk}{0.65,0,0.1,0.3}
\definecolor{purple}{rgb}{0.65,0,0.65}
\definecolor{dark_green}{rgb}{0, 0.5, 0}
\definecolor{orange}{rgb}{0.8, 0.6, 0.2}
\definecolor{red}{rgb}{0.8, 0.2, 0.2}
\definecolor{darkred}{rgb}{0.6, 0.1, 0.05}
\definecolor{blueish}{rgb}{0.0, 0.3, .6}
\definecolor{light_gray}{rgb}{0.7, 0.7, .7}
\definecolor{pink}{rgb}{1, 0, 1}
\definecolor{greyblue}{rgb}{0.25, 0.25, 1}

\def \customparskip {.5em}
\renewcommand{\paragraph}[1]{\vspace{\customparskip}\noindent\textbf{#1}.}

\begin{document}
\pagestyle{headings}
\mainmatter

\def\ECCVSubNumber{5280}  %

\title{Estimating Visual Information From Audio Through Manifold Learning} %

\titlerunning{Estimating Visual Information from Audio}
\author{%
Fabrizio Pedersoli\inst{1}\and%
Dryden Wiebe\inst{1}\and%
Amin Banitalebi \inst{2}\and%
Yong Zhang\inst{2}\and
George Tzanetakis\inst{3}\and
Kwang M. Yi\inst{1}%
}
\authorrunning{F. Pedersoli et al.}
\institute{
University of British Columbia \email{\{fpeder,drydenw, kmyi\}@cs.ubc.ca}\and
Huawei Technologies Canada Co., Ltd \email{\{amin.banitalebi,yong.zhang3\}@huawei.com}\and
University of Victoria \email{gtzan@uvic.ca}%
}

\maketitle

\begin{abstract}

We propose a new framework for extracting visual information about a scene only using audio signals. 
Audio-based methods can overcome some of the limitations of vision-based methods i.e., they do not require ``line-of-sight'', are robust to occlusions and changes in illumination, and can function as a backup in case vision/lidar sensors fail. Therefore, audio-based methods can be useful even for applications in which only visual information is of interest.
Our framework is based on Manifold Learning and consists of two steps.
First, we train a Vector-Quantized Variational Auto-Encoder~\cite{van2017neural} to learn the data manifold of the  particular visual modality we are interested in.
Second, we train an Audio Transformation network to map multi-channel audio signals to the latent representation of the  corresponding visual sample.
We show that our method is able to produce meaningful images from audio using a publicly available audio/visual dataset.
In particular, we consider the prediction of the following visual modalities from audio: depth and semantic segmentation. We hope the findings of our work can facilitate further research in visual information extraction from audio. Code is available at: \url{https://github.com/ubc-vision/audio_manifold}.


\end{abstract}

\section{Introduction}
\label{sec:intro}

\emph{Line-of-Sight} is a fundamental requirement of many computer vision algorithms and applications~\cite{redmon2016you,ren2015faster,chen2018encoder,janai2020computer}. Most algorithms must have ``clear'' vision of the entire scene in order to properly work.
If this requirement is not met, the algorithms' effectiveness is drastically compromised to a point they can not be used at all. 
Computer vision tasks such as object localization~\cite{tompson2015efficient,caicedo2015active,choe2020evaluating,zhang2018adversarial} and autonomous navigation~\cite{royer2005outdoor,royer2007monocular,scaramuzza2014vision}
are particularly affected by the line-of-sight problem.
Moreover, when it comes to indoor vision applications~\cite{fusco2020indoor,sun2019see,baek2019augmented,kumar2014accurate,sadeghi2014weighted}, meeting the line-of-sight requirement becomes even more challenging. 
In indoor situations occluding objects are more likely to be encountered and the layout of the environment can make tasks such as localization or navigation very difficult due to limited vision of the entire scene.
On the other hand, potential \emph{audio-based} methods do not require a direct line-of-sight\footnote{Audio based application are also less \emph{privacy} concerning.
While it is true that audio carries conversation, this can be removed from the audio signal by using sound source separation algorithms.}.
For instance, the sound of a dog barking is a reasonable enough cue to infer that a dog is nearby even if it is not visible.
Prior works have shown that \emph{sounding objects} can be localized in space with audio-only techniques~\cite{willert2006probabilistic,gerstoft2021audio,valin2003robust} by leveraging multi-channel processing using microphone arrays.
More recent research works on audio have shown that even more involved audio-based methods can be designed, such as methods that estimate rough visual characteristics of a scene~\cite{christensen2020batvision,vasudevan2020semantic} in terms of depth and semantic segmentation.
Such audio techniques for extracting visual information are limited to the detection of sounding objects, or require complex instrumentation for estimating the characteristics of indoor scenes.

In this paper we propose a method for extracting visual information from audio which is not limited to sounding objects and does not require any ad-hoc instrumentation. 
Specifically, we propose a method capable of predicting: %
depth, and semantic segmentation of a scene.
In order to achieve this goal, we propose a novel deep learning framework based on \emph{manifold learning}.
Given an audio/video pair\footnote{An audio/video pair consists of a small segment of audio, typically $1$ or $2$ seconds, and its corresponding frame of the desired visual modality.}, our method learns to \emph{transform} an audio segment to its corresponding frame visual modality by using their encoded (latent) representation.
At first, a Vector-Quantized Variational Auto-Encoder (VQ-VAE)~\cite{van2017neural} is trained on the target visual modality in order to learn the manifold data.
This process is iterated for all the visual modalities.
Note that we do not use the VQ-VAE for generating new data.
Once the data manifold of the visual modality is learned, we train the \emph{Audio Transformation} network (AT-net).
The purpose of AT-net is to encode an audio segment into a latent representation which is ``close'', in terms of $L_2$ norm, to the corresponding visual sample in its latent space (not quantized). 
The transformed latent code is then quantized and passed through the pretrained visual decoder to get the final visual reconstruction.
We evaluate our method on two publicly available audio/video datasets both for outdoor urban scenes. 

Our contributions are summarized as follows: $(i)$ We propose new manifold learning framework for estimating visual information from audio which can produce accurate depth maps and semantic segmentation maps. 
$(ii)$ We show that VQ-VAE are a powerful tool for learning a manifold for this task.
$(iii)$ We define an audio transformation network architecture which can effectively learn the transformation between audio and visual sample at the manifold level.

\section{Related work}
\label{sec:related}

In this section we describe previous work on audio and audio-vision algorithms for estimating visual information.
In the first part we describe audio based methods for sound source localization. Then, we overview previous work which uses audio and vision for estimating visual cues. Finally, we describe more recent work on audio only techniques for estimating visual information.

\paragraph{Audio sound source localization}
Earlier research focused on the problem of Sound Source Localization (SSL) using microphone arrays which often have more than two elements.
The main idea for solving SSL is to exploit the different times of arrival of sound in each microphone, as well as the different gain values, to regress the spatial position of the sound source.

For instance, Willert et al.~\cite{willert2006probabilistic} compute activity maps based on cochlear modelling of binaural audio. The activity maps are frequency/time-delay representations which are used by a probabilistic model to estimate the position (azimuth) given as input reference activity maps at known positions.
Gerstoft et al.~\cite{gerstoft2021audio} propose a system for localizing sound sources by using 5-8 circular arrays of microphones. Direction of arrival was computed for each array and then fed to a processing method, which is either based on PCA or affine transformation. 
Valin et al.~\cite{valin2003robust} present a method for sound source localization in 3D space using an array of 8 microphones. Their method is based on time delay of arrival estimation and can work in real-time.
Otsuka et al.~\cite{otsuka2013bayesian} present a unified framework for sound source localization and source separation which is jointly optimized and based on bayesian nonparametrics.
Yalta et al.~\cite{yalta2017sound} propose the use of deep neural network to localize sound sources using a microphone array.

All the aforementioned previous work need to operated in controlled environments, and can only estimate the position of the sounding object as coordinate in space. In our work we aim at localization objects as ``masks'' in 2D images that can reproduce some visual characteristic of the object.

\paragraph{Audio-visual learning}
More recent work has used audio jointly with vision in order to identify peculiar regions within the image based on sound characteristics.
For example, one task is the localization of sound sources in images in terms of heat-maps.

In that regard, Senocak et al.~\cite{senocak2018learning} propose a two stream Convolutional Neural Network (CNN) to perform sound localization in an unsupervised manner. The output of CNN is processed by an attention module which combines the audio and vision stream. Their method can also work in supervised/semi-supervised settings.
Aytar et al.~\cite{aytar2016soundnet} present a deep architecture for learning sound representations from unlabelled data.
The inherent correlation of sound and vision in a video, allowed them to train in a student teacher setting a model that yields superior performance for acoustic/object classification. 
Arandjelovi\'{c} and Zisserman~\cite{arandjelovic2018objects} propose a method for locating sounding objects in images given the audio signal.
In particular, they use a two stream network which is trained with the objective of audio and video correspondence.

In a similar way, by leveraging massive amounts of unlabelled data in unsupervised learning setting, Zhao et al.~\cite{zhao2018sound} present a method for locating sounding regions within an image. Their method is jointly audio-visual and is capable of separating sound sources, and mapping them to their corresponding pixel locations.
As an extension of their previous work, Zhao et al.~\cite{zhao2019sound}, inspired by the fact that sound is generated by the motion of objects and vibrations, propose a deep learning system capable of extracting motion cues for the task of sound source localization and separation.
By using unlabelled audio/video data the authors train end-to-end a deep model in a curriculum learning setting.
Their model is composed of: motion network, appearance network, and sound source separation network; where audio and visual streams are interconnected with an attention-like module.

These works, use audio as an additional source of information for detecting peculiar regions with the image based on characteristics of sound. 
In our work, instead, we aim to estimate visual information from audio only.

\paragraph{Estimating vision from audio}
A new emerging and challenging research direction is to develop methods capable of extracting visual information from audio only. 
Inspired by echolocation in animals, Christensen et al.~\cite{christensen2020batvision} designed a system for estimating depth maps and grayscale images of indoor scenes using a radar-like approach.
Their system is composed of a binaural microphone, a speaker, and a ZED camera mounted on a small moving agent.
The agent emits small chirp signals and the returning echos are modelled by a deep network to reconstruct depth or grayscale images using the camera image as ground-truth.

Ire et al.~\cite{irie2019seeing} propose a method for predicting semantic segmentation of human bodies in a controlled environment from multichannel audio.
Angular spectrum and MFCCs (Mel Frequency Cepstrum Coefficient) are extracted from the audio and fed to a two stream convolutional encoder.
After a fusion layer, a convolutional decoder is used for predicting the segmentation map.
Lin et al.~\cite{lin2021unsupervised} present an unsupervised method for identifying sounding pixels in images by using a contrastive learning framework.
Their method is iterative, at first they learn the correlation between audio and visual signals (within the same audio/visual pair) which are then used as pseudo-labels for the subsequent step.
In the next step they learn the audio/visual correspondence across different videos which allows to refine the detection of sounding pixels. 

Valverde et al.~\cite{valverde2021there} propose a self supervised framework for detecting cars bounding boxes from sound by using a student-teacher approach.
A student-teacher approach is also used in the work of Vasudevan et al.~\cite{vasudevan2020semantic}.
By using an encoder-decoder architecture they predict segmentation maps of sounding objects (car, train, motorcycle), as well as, depth maps, and audio super-resolution.

The Previously described works for estimating visual information has some limitations we are interested to overcome with our proposed method. These limitation are either: to required complex setup and instrumentation, or considering the estimation of tiny portion of what it would be entire visual scene.

\section{Method}
\label{sec:method}
\begin{figure}
    \centering
    \includegraphics[width=.99\textwidth]{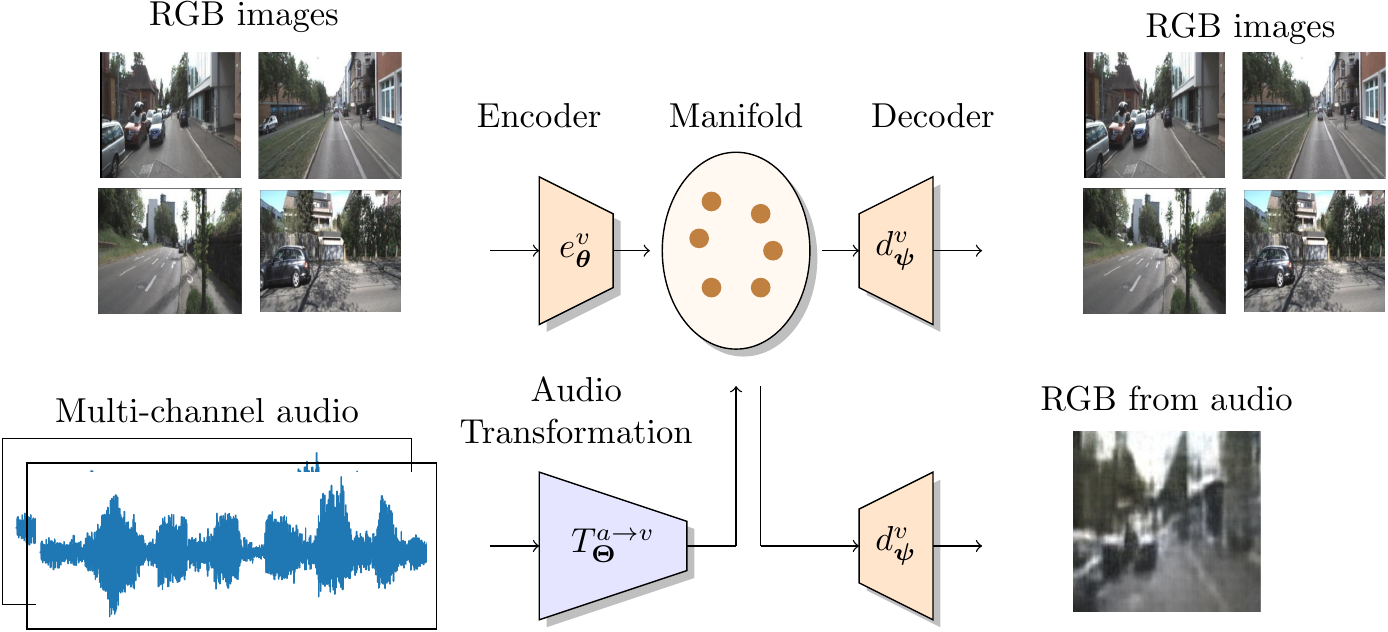}
    \caption{Overview of our method. At first stage the data manifold is learnt for each visual modality by using a VQ-VAE. After that, the audio transformation network maps an audio sample to the closet visual sample in manifold space.}
    \label{fig:method}
\end{figure}

In this paper we propose a two-stage method for extracting visual information from audio, as shown in Fig.~\ref{fig:method}. 
The first stage consists of training a VQ-VAE on a particular visual modality, which can be but not limited to: depth maps, and semantic segmentation maps. The second stage consists of training a domain transformation network, which we refer to as Audio-Transformation network (AT-net), to map audio signals to the visual modality.

We propose this two-stage approach because it provides advantages compared to single-stage (end-to-end) approaches. End-to-end models have limitations when used to extract visual information from audio signals. We have empirically observed that such models converge to an average representation of the data-set that has low quality and lacks visual detail. 
A two-stage approach can overcome these issues by \emph{learning a transformation at the manifold}\footnote{We refer to manifold as the space defined by the encoded representation of some data type (visual data in our case).} level. 
The key idea of learning a transformation at the manifold level is that it potentially allows to reconstruct the overall structure, as well as the details, of a visual modality because the decoder is specific to that modality. This transformation can be effectively learned because manifold data lies on an higher dimensional space which enforce sparsity and thus is less affected by the ``regression to the mean'' representation problem. 

\subsection{VQ-VAE Pretraining}
\label{ssec:vqvae}
\begin{figure}[t]
    \centering
    \includegraphics[]{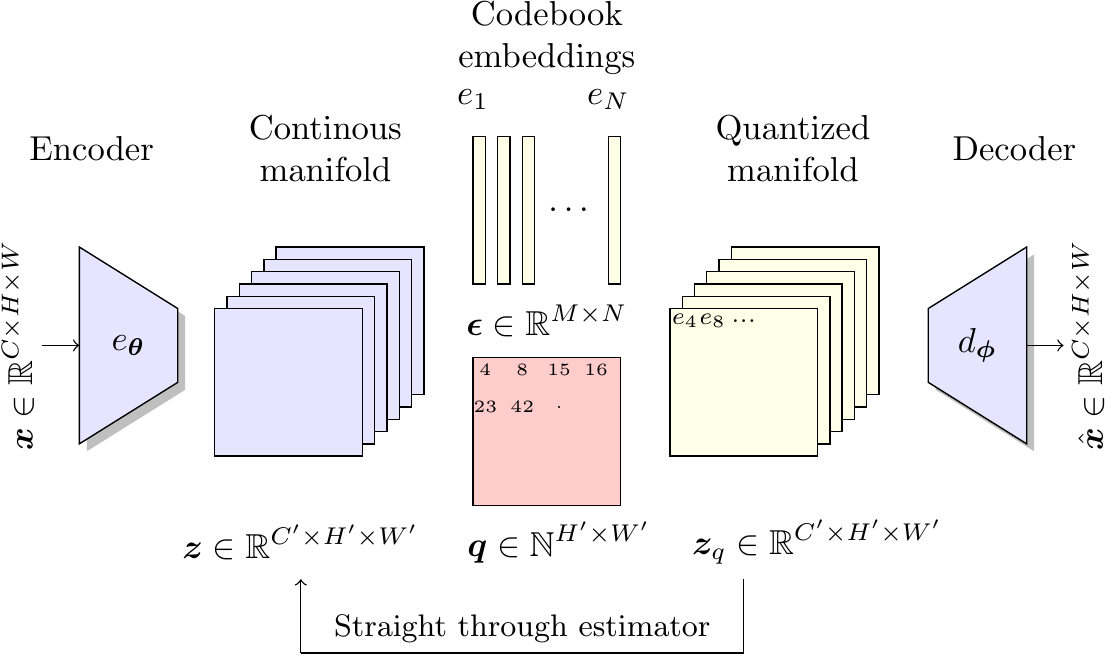}
    \caption{Vector-Quantized Variational Auto Encoder framework.}
    \label{fig:vqvae}
\end{figure}

We use a VQ-VAE~\cite{van2017neural} framework for learning the manifold of a given visual modality, in our case depth, and segmentation maps. 
See Fig.~\ref{fig:vqvae} for an overview of the VQ-VAE framework.
The main idea behind the VQ-VAE is to introduce a trainable vector quantizer at a latent space (manifold).
The model is trained with reconstruction loss plus two loss terms specific for the vector quantization. 
During back-propagation the gradient is simply passed through by a ``straight through estimator'' \cite{bengio2013estimating} to the vector quantization module.
Deep learning research has shown that VQ-VAEs do not have any particular limitation compared to VAEs \cite{kingma2013auto}, while simplifying a lot the latent representation (manifold) of the data samples.

In our case, the VQ-VAE allows us to obtain a quantized representation of latent data which is more ``controllable''. In our experiments we can get very accurate reconstruction just by using a small number of learnable embeddings (the code-book), e.g $N=64$. This is possible because we independetly train a specific VQ-VAE for each modality. In our experience we note that the principal factor for obtaining detailed reconstruction is the higher resolution of the latent space\footnote{In our experiments a latent size $32\times 32$ allows us to reconstruct data at $128\times 128$ size with no visible loss of details.}. On the other hand, smaller resolutions of the latent space yield progressively more blurry reconstructions.
It is important to note that the latent space resolution is a fundamental design choice of the overall method. In fact, the AT-net entirely relies on this data for learning to map audio to visual information; the resolution size determines how effectively this transformation can be learned.
Learning this transformation is not an easy task, and we specifically rely on the VQ-VAE for making this process more effective. In fact, the quantized nature of the framework allows the model to better tolerate small ``errors'' in transformation since they will be eliminated by the quantization process. Finally, note that we are not interested in any sort of data generation (sampling). 

\subsection{Audio Transformation}
\label{ssec:audio-tran}
The purpose of the AT-net is to encode an audio sample to the closest  manifold sample in the visual domain; where closest is defined in terms of the $L_2$ norm.
Audio transformation is a fundamental processing step for obtaining the good overall performance of the proposed method. Therefore, the architecture of AT-net must be carefully chosen.

\begin{figure}[t]
    \centering
    \includegraphics{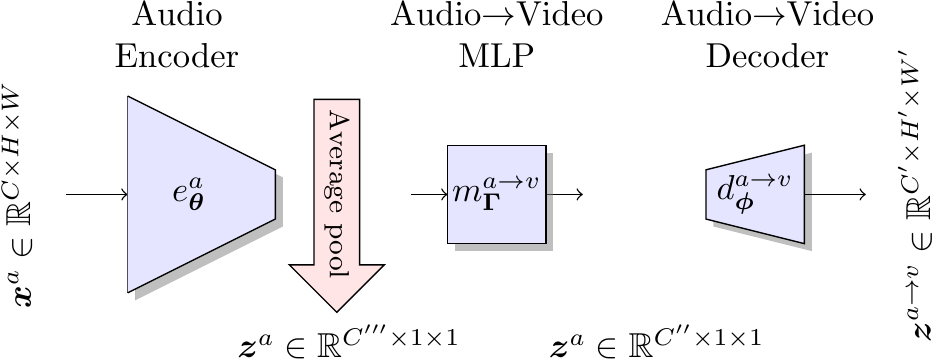}
    \caption{The Audio transformation network.}
    \label{fig:at-net}
\end{figure}

The AT-net consists of three components: audio encoder, domain transformation MLP, and visual manifold decoder, as shown in Fig.~\ref{fig:at-net}.
The audio encoder consisting of a series of strided convolutional layers which considerably reduce the size of data. A final global average pooling layer reduces the spatial size to $1\times 1$. Note that completely reducing the spatial dimension of the data is important because there is no spatial correspondence between the two domains.
The domain transformation MLP consists of several dense layers which are used for decoupling the audio and visual domains.
Finally, the visual decoder increases the size of data to match the chosen visual manifold size by means of strided transposed convolutional layers.

The AT-net is simply trained with $L_2$ loss at the manifold level. In more detail, given a training (audio, visual) pair, at first the visual sample must be encoded on the visual manifold by using the VQ-VAE encoder. Then, this manifold data sample acts as ground truth for training the AT-net.
Note that we train the AT-net on the continuous (not quantized) latent representation of the visual data. We do this because we want quantization to be the subsequent step as it can remove errors in transformation.
As already mentioned in the previous subsection, the resolution of the manifold is a key design factor.
A small manifold resolution yields easier AT-net training but more blurry reconstruction.
On the other hand, a bigger manifold resolution ensures detailed reconstruction, but the training process of AT-net is more problematic.

\section{Results}
\label{sec:results}

In this section we report the experimental evaluation of our method with respect to the prediction of depth maps and segmentation maps from audio signal.
We evaluate our method on a publicly available datasets: the ``Omni Audiotory Perception'' (OAP) \cite{vasudevan2020semantic}.
This dataset provides multi-channel audio and video of 165 urban scenes recorded by equipment mounted on a car.
Data is recorded on different days and locations.
In particular, audio is $8$ channels, e.g. $8$ microphones are used to record the environment sounds. 
Video comes in the form of high-resolution RGB frames and depth maps. 
Depth maps are computed from single RGB frames by using the official implementation of \texttt{Monodepth2} \cite{godard2019digging} which estimates depth maps from monocular frames.
In addition to that, we also compute the semantic segmentation maps by using an implementation of \texttt{Deeplabv3} architecture \cite{chen2017rethinking} pretrained on the ``Cityscapes'' datasets \cite{cordts2016cityscapes}. 

\paragraph{Data preparation}
We resample audio to \SI{22.5}{\kilo\hertz}. 
By doing this we obtain a more compact representation, i.e more computationally efficient, which does not lose any meaningful information of sound.
Following the same methodology as \cite{vasudevan2020semantic,valverde2021there} 
we process audio/video in windows of \SI{1}{\second}. 
Therefore, an audio/video pair consists of: \SI{1}{\second} of audio and the corresponding central video frame of the same temporal window.
As a final pre-processing step we compute the MEL spectrogram of \SI{1}{\second} audio segments. 
The choice of MEL spectrogram is motivated by previous research in audio (music) that showed the effectiveness of this time-frequency representation for many downstream tasks \cite{logan2000mel}.
We compute spectrograms according the following specification: window length of $2048$ samples ($L$) (same as \#FFT points), hop length of $256$ samples ($H$), and number of mel coefficients $256$ ($W$).
Note that we use all the available audio channels ($N$) by concatenating the respective spectrograms along a new dimension such that the final input for the audio transformation network is $N\times \lfloor\frac{L}{H} \rfloor \times W$ .
The images of the different visual modalities are simply resized to our input resolution of $128\times 128$ . Spectrograms are resized as well to the same size.

\paragraph{Networks specifics} 
For the VQ-VAE we adopt a similar configuration as the one proposed by the original authors \cite{van2017neural}. 
The encoder and decoder are characterized by simple strided convolutional layers with ReLU activations.
The number of features for the first convolutional layer, both at the encoder side and decoder side, is set to $64$.
All the remaining convolutional layers have $128$ features. Note that each convolutional layer is \emph{strided}, i.e. it downsamples (encoder) or upsamples (decoder) its input by a factor of $2$. 
A sequence of three residual convolutional blocks \cite{he2016deep} is placed before and after the vector quantization module; the residual blocks do not change their input size.
With respect to the vector quantizer, we configure it to have $64$  code-words of dimensionality $64$.
The number of strided convolutions of the encoder and decoder is chosen in order to match the desired size of the data manifold: $8\times 8$, $16\times 16$, and $32\times 32$ in our experiments.

The other architectural component of our method is the AT-net.
As previously described in Sec.~\ref{ssec:audio-tran}, the AT-net consists of three parts: audio encoder, domain transformation MLP, and visual manifold decoder.
Specifically, the audio encoder is a Resnet18 \cite{he2016deep}.
The domain transformation MLP consists of a sequence of three dense layers with dropout probability of $p=0.2$ and ReLU activations.
The visual decoder consists of series of strided transposed convolutional layers which upsamples at each stage the output of the MLP by a factor of $2$ until the desired manifold size is reached; the number of features is halved at each stage. 
As final note, we add batch normalization~\cite{ioffe2015batch} to increase training speed and stability.

\paragraph{Training} 
We train our method using the same data splits as suggested by the datasets authors. 
We use the Adam optimizer~\cite{kingma2014adam} with learning rate of \num{1e-4} for both the VQ-VAE and AT-net training. We stop training when the incremental improvement of the training loss observed over a fixed amount of iterations does not improve over a certain threshold empirically chosen. 

\paragraph{Evaluation}
In regard to depth estimation, we conduct the evaluation by using several errors metrics as defined in Eigen et al.~\cite{eigen2014depth}.
Specifically, we use: absolute relative distance ($ABS_{\text{rel}}$), squared relative distance ($SQR_{\text{rel}}$), RMSE linear ($RMSE_{\text{lin}}$), and RMSE logarithmic ($RMSE_{\text{log}}$). In addition, we also evaluate $AUC_{\text{crr}}$ of relative correct detections below thresholds $\tau \in \{0,0.01,\ldots 0.3\}$.
By using the notation: $y$ ground truth image, $\hat{y}$ predicted image, $T$ number of data samples, $N$ number of pixels of an image, the evaluation metrics are defined as follows.
\begin{equation}
\begin{gathered}
ABS_{\text{rel}} = \frac{1}{|T|}\sum_{y\in T}|y - \hat{y}|/\hat{y}\,, \quad
SQR_{\text{rel}} = \frac{1}{|T|}\sum_{y\in T}\|y - \hat{y}\|^2 / \hat{y}\,, \\
RMSE_{\text{lin}} = \sqrt{\frac{1}{|T|}\sum_{y\in T}\left\|y - \hat{y}\right\|^2}\,, \quad
RMSE_{\text{log}} = \sqrt{\frac{1}{|T|}\sum_{y\in T}\|\log{y} - \log{\hat{y}}\|^2} \\
AUC_{\text{crr}} = \sum_{t \in |\tau|} Crr_{t} \cdot (\tau_{t+1} - \tau_{t}) \,,\quad
Crr_{t} = \frac{1}{|N|} \sum_{i\in N}\mathbbm{1}\!\left[
\frac{| y_i - \hat{y_i}|}{y_i} \right]_{<\tau_t}.
\end{gathered}
\end{equation}

In addition, for the semantic segmentation evaluation we report performance results in terms of mean Intersection Over Union (mIOU), alongside the individual IOU of each one of ground truth classes. Classes that scores an IOU of below $<1\%$ for all the considered methods are removed from the evaluation.

\paragraph{Consideration about metrics}
It is important to note that the used evaluation metrics do not completely take into account some key aspects that define a prediction to be ``good'' or ``bad'', as a human would do. 
The used metrics are averaged pixel level metrics.
Thus, they do not consider the overall ``structural fit'' of the prediction, as well as higher level semantics such as ``correspondence of objects''. In fact, a prefect prediction affected by small translation or deformation would still end up considered ``bad'' from these metrics point of view.
While still providing  valid comparison with other methods, these metrics are not ideal for evaluating methods which does not take preservation of the structure and objects semantics for granted\footnote{The methods discussed in this section predict visual information from audio. This tasks is really challenging and predicting something which has a structural fit is already an important achievement.}. 
Ideally, a new metric that takes into account the aforementioned aspects should be defined for this task. However, this is outside the scope of this paper. To compensate what the metrics are missing, we qualitatively show some predicted examples of our method and compare them with previous work.

\paragraph{Comparison}
We compare our method with the method proposed by Vasudevan et al.~\cite{vasudevan2020semantic} by using the implementation available at their github repository.
In order to conduct a fair evaluation we test their method by using both 2ch and 8ch audio input.
In addition, since their methods obtains the best performance when predicting multiple visual sources, we also include these configurations in our experimental evaluation. However, we do not include the super-resolution audio prediction. We also include an additional baseline comparison with ECHO2DEPTH \cite{gao2020visualechoes} for depth estimation. 

For our method we report performance with respect to different spatial resolutions of the data manifold, namely: $8\times 8$, $16\times 16$, and $32\times 32$.
For both the cases of visual estimation, our predicted images are at a smaller resolution with respect to the compared method, therefore, for a fair evaluation we resize the predicted images with nearest neighbour interpolation to match the target resolution.
To summarize, we compare our method with the following configurations of method \cite{vasudevan2020semantic}: 2ch audio $\rightarrow$ depth, 2ch audio $\rightarrow$ segmentation, 2ch audio$\rightarrow$ depth $+$ segmentation, 8ch audio $\rightarrow$ depth, 8ch audio $\rightarrow$ segmentation, and 8ch audio $\rightarrow$ depth $+$ segmentation. 

\subsection{Omni auditory perception dataset}
\label{ssec:oap}

In Tab.~\ref{tab:oap_depth} we report the performance comparison for the task of depth maps estimation.
\begin{table}[t]
    \caption{OAP depth map estimation results.}
    \label{tab:oap_depth}
    \begin{center}
    \setlength{\tabcolsep}{3pt}
\begin{tabular}{@{}lcccccc@{}}
\toprule
Method & $ABS_{\text{rel}}\downarrow$ & $SQR_{\text{rel}}\downarrow$  & $RMSE_{\text{lin}}\downarrow$ & $RMSE_{\text{log}}\downarrow$ & $AUC_{\text{crr}}\uparrow$  \\
\midrule
\cite{gao2020visualechoes} ECHO2DEPTH       &  .731     & 6.13     & 7.00   & 1.19  & .077   \\
\cite{vasudevan2020semantic} $2$ch       &  .478     & 2.78     & 5.12   & .527  & .068   \\
\cite{vasudevan2020semantic} $8$ch       &  .674     & 4.51     & 5.29   & .602  & .047  \\
\cite{vasudevan2020semantic} $2$ch +seg  &  .403  & 1.73 & 4.54  & .474  & .077   \\
\cite{vasudevan2020semantic} $8$ch +seg  &  .629  & 3.16 & 5.25  & .596  & .053  \\
\midrule
Proposed $8\times 8$                  &  \bfseries .241  & \bfseries 1.01 & 4.12  & \bfseries .361  & \bfseries .118  \\
Proposed $16\times 16$                &  .235  & 1.04 & \bfseries 4.11  & .376 & .118  \\
Proposed $32\times 32$                &  .251  & 1.08 & 4.16  & .382 & .114  \\
\bottomrule
\end{tabular}
 
    \end{center}
\end{table}

As we can see from Tab.~\ref{tab:oap_depth} our method obtains the best performance for all the performance metrics. 
Previous method \cite{vasudevan2020semantic} performs at best in 2ch audio multi modal configuration, i.e. when both depth maps and semantic segmentation maps are predicted. However, we operating in single visual modality the performance of their method are heavily penalized. 
Focusing on our method, the $16\times 16$ configuration is the best performing among all our configurations.
The manifold at $16\times 16$ resolution is particularly effective for depth estimation because the visual data is already smooth and slowly variant. Therefore, the advantages offered by higher manifold resolutions for reconstructing high-frequency details are not as important in this particular case. Thus, our model can be effectively trained and perform best at the lowest manifold resolution.

In Tab.~\ref{tab:oap_seg} we report the performance comparison for the task of semantic segmentation maps estimation. For this experiment we consider the IOU of classes which are above the threshold of $0.01\%$.
\begin{table}[t]
    \caption{OAP semantic segmentation results reported in IOU [\%]. %
    }
    \label{tab:oap_seg}
    \begin{center}
    \adjustbox{max width=\textwidth}{%
\setlength{\tabcolsep}{3pt}
\begin{tabular}{@{}l|r|rrrrrrrr@{}}
\toprule
Method & $AV\!G$ & Road & Side. & Build. & Fence & Veget. &  Terr. & Sky & Car \\
\midrule
\cite{vasudevan2020semantic} 8ch        & 17.66  & 74.94 & 0.47 & 27.35 & 0.29 & 14.05 & 0.06 & 18.69  & 5.48   \\
\cite{vasudevan2020semantic} 2ch        & 18.83  & \bfseries81.90 & 0.58 & 26.63 & 0.09 & 10.88 & 0.03 & 24.94  & 5.59   \\
\cite{vasudevan2020semantic} 8ch +depth & 18.30  & 79.96 & 0.70 & \bfseries 29.02 & 0.19 & 9.49 & 0.02 & 20.48  & 6.21   \\
\cite{vasudevan2020semantic} 2ch +depth & 19.20  & 76.77 & 0.33 & 25.68 & 0.21 & \bfseries 14.51 & 0.22 & 28.27  & \bfseries 7.47   \\
\midrule
Proposed $8\times 8$ & 20.73 & 74.67 & \bfseries 4.34 & 11.79 & 1.53 & 15.11 & \bfseries2.59 & 53.31  & 2.49   \\
Proposed $16\times 16$ & \bfseries20.79 & 75.31 & 4.08 & 12.83 & \bfseries 1.98 & 14.53 & 2.44 & 52.54  & 2.63   \\
Proposed $32 \times 32$ & 20.29 & 73.63 & 4.14 & 11.13 & 1.20  & 12.51 & 1.18 & \bfseries55.60  & 2.94   \\
\bottomrule
\end{tabular}
}

    \end{center}
\end{table}
By looking at the average IOU we see that our method is clearly the best performing method. The margin of improvement offered by our method is considerably large, almost double the performance score of \cite{vasudevan2020semantic} method.
In the case of semantic segmentation estimation, their method performs similarly for both the single and multi-modal configuration;
taking into account the depth modality did not provide any sensible improvement. 
Our method performs similarly among the three different manifold resolutions. As happens for the depth data, also in this case due to the segmentation map is slowly variant in regions which often comprise a large number of pixels.  

Focusing on the individual semantic segmentation classes, we see that both ours and their method perform very well for class `Road' scoring $\approx 75\%$ IOU. This could be due to scene characteristics that are encountered in both training and testing data. For class `Sky' instead, our method provides a huge advantage with respect to \cite{vasudevan2020semantic}. In fact, for this class we obtain an IOU of $\approx 53\%$, while the compared method only scores $\approx 30\%$ in their best configuration. Conversely, for class `Building' and class `Car' their method perform better than ours. 
A possible explanation for this is the loss of details due to small manifold resolutions. Due to the nature of the adopted VQ-VAE framework when dealing with segmentation data, possible small regions of a particular class which are adjacent with much larger regions of another class can be lost during decoding.
Moreover, we also not that our method is able to detect some regions while the compared method almost misses those. In particular, although with a low IOU our method can identify classes `Sidewalk', `Fence', and `Terrain'.

\subsection{Multimodal audio visual detection dataset}
\label{ssec:mavd}

In Tab.~\ref{tab:mavd_depth} we report the performance comparison for the task of depth maps estimation.
\begin{table}[t]
    \caption{MAVD depth map estimation results.}
    \label{tab:mavd_depth}
    \begin{center}
    \setlength{\tabcolsep}{3pt}
\begin{tabular}{@{}lcccccc@{}}
\toprule
Method & $ABS_{\text{rel}}\downarrow$ & $SQR_{\text{rel}}\downarrow$  & $RMSE_{\text{lin}}\downarrow$ & $RMSE_{\text{log}}\downarrow$ & $AUC_{\text{crr}}\uparrow$  \\
\midrule
\cite{gao2020visualechoes} ECHO2DEPTH     &  .218  & .561 & 2.34  & .324  & .14  \\
\cite{vasudevan2020semantic} $2$ch       &  .232  & .740 & 2.69  & .324  & .37  \\
\cite{vasudevan2020semantic} $8$ch       &  .210  & .669 & 2.54  & .301  & .46  \\
\cite{vasudevan2020semantic} $2$ch +seg  &  .223  & .676 & 2.55  & .313  & .46  \\
\cite{vasudevan2020semantic} $8$ch +seg  &  .205  & .634 & 2.49  & .298  & .148  \\
\midrule
Proposed $8 \times 8$   &  \bfseries .126 & \bfseries .290 & \bfseries 1.51  &\bfseries .180 & \bfseries .191  \\
Proposed $16 \times 16$ &  .146 & .347 & 1.65  & .201 & .184  \\
Proposed $32 \times 32$ &  .139 & .303 & 1.65  & .196 & .187  \\
\bottomrule
\end{tabular}
 
    \end{center}
\end{table}
As we can see from the table, our method at manifold resolution of $8\times 8$ achieves the best performance according to all the metrics by a considerable margin with respect to the baseline methods. For the bigger manifold resolution the performance decrease but still the proposed method outperforms the baseline method.

In Tab.~\ref{tab:mavd_seg} we report the performance comparison for the task of   semantic segmentation maps estimation.
\begin{table}[t]
    \caption{MAVD semantic segmentation results reported in IOU [\%]. %
    }
    \label{tab:mavd_seg}
    \begin{center}
    \adjustbox{max width=\textwidth}{%
\setlength{\tabcolsep}{3pt}
\begin{tabular}{@{}l|r|rrrrrrrr@{}}
\toprule
Method & $AVG$ & Road & Side. & Build. & Fence & Veget. &  Terr. & Sky & Car \\
\midrule

\cite{vasudevan2020semantic} 2ch        & 36.66  & 76.42 & 25.59 & 48.56 & 16.16 & 58.88 & 14.86 & 26.47  & 26.40   \\
\cite{vasudevan2020semantic} 8ch        & 40.05  & 78.75 & 28.96 & 52.68 & 18.65 & 63.56 & 17.59 & 29.81  & 30.47   \\
\cite{vasudevan2020semantic} 2ch +depth & 41.68  & 76.12 & 29.10 & 49.18 & 26.86 & 59.92 & 25.68 & 39.70  & 26.94   \\
\cite{vasudevan2020semantic} 8ch +depth & 47.36  & 80.15 & 34.83 & 55.31 & 30.71 & 64.46 & 31.61 & 47.74  & 34.11   \\

\midrule

Proposed $8\times 8$     & \underline{\textbf{61.02}} & \textbf{90.42} & \textbf{55.43} & \textbf{77.61} & \textbf{38.23} & \textbf{84.56} & \textbf{39.68} & \textbf{47.64}  & \textbf{54.62}   \\
Proposed $16\times 16$   & 57.99 & 89.18 & 49.78 & 75.98 & 35.00 & 82.76 & 36.06 & 45.90  & 49.28   \\
Proposed $32\times 32$   & 54.83 & 87.99 & 45.88 & 73.52 & 30.70 & 81.13 & 31.12 & 43.90  & 44.47   \\
\bottomrule
\end{tabular}
}

    \end{center}
\end{table}
Also for this task we note the proposed method obtains the best performance for all of the semantic segmentation classes. Similarly to the previous task, the best configuration of our model is at manifold resolution of $8\times 8$. For the bigger manifold resolution the performance slightly decrease.

\subsection{Ablation study}
In this section we discuss the ablation study for VQ-VAE and VAE as framework for learning the data manifold. With this experiments we aim at showing which one of the two frameworks allows to learn are more suitable manifold for estimating visual information. In addition, we also report the performance of our method trained end-to-end (E2E) in a single stage i.e. manifold and audio transformation is learnt at the same time.

In Tab.~\ref{tab:vqnvq_depth} we report the comparison with respect to the prediction of depth maps.
\begin{table}[t]
    \caption{VQ-VAE vs VAE vs E2E results for depth maps estimation.}
    \label{tab:vqnvq_depth}
\begin{center}
    \setlength{\tabcolsep}{3pt}
\begin{tabular}{@{}lcccccc@{}}
\toprule
Method & $ABS_{\text{rel}}\downarrow$ & $SQR_{\text{rel}}\downarrow$  & $RMSE_{\text{lin}}\downarrow$ & $RMSE_{\text{log}}\downarrow$ & $AUC_{\text{crr}}\uparrow$  \\
\midrule
E2E $8\times 8$        &  .442  & 2.31 & 4.92  & .526  & .072  \\
E2E $16\times 16$      &  .496  & 2.25 & 4.79  & .529  & .068  \\
E2E $32\times 32$      &  .541  & 3.16 & 5.15  & .561  & .062  \\
VAE $8\times 8$        &  .311  & 1.22 & 4.11  & .397  & .097  \\
VAE $16\times 16$      &  .261  & 1.15 & 4.07  & .377  & .112  \\
VAE $32\times 32$      &  .248  & 1.10 & \bfseries 4.01  & .381  & .113  \\
\midrule
VQ $8\times 8$         &  \bfseries .241  & \bfseries 1.01 & 4.12  & \bfseries .361  & \bfseries.118  \\
VQ $16\times 16$       &  .235  & 1.04 & 4.11  & .376  & .118  \\
VQ $32\times 32$       &  .251  & 1.08 & 4.16  & .382  & .114  \\

\bottomrule
\end{tabular}
\end{center}
\end{table}
From Tab.~\ref{tab:vqnvq_depth} we see that VQ-VAE are indeed the better framework for learning the data manifold. In fact, for all of the resolutions of manifold data, VQ-VAEs outperform VAEs counterpart for all the performance metrics. Interestingly, we also note that in the case of VAE the better (in the sense of majority of the performance metrics) manifold resolution is $32\times 32$ while for VQ-VAE the better manifold resolution is $8\times 8$. When the proposed model is trained end-to-end we note that the performance are worse with respect to the two-stage models.

\subsection{Qualitative results}

In this section we show some qualitative results of depth map estimation and semantic segmentation estimation from audio. We compare our $8 \times 8$ method with the method proposed by Vasudevan et al.~\cite{vasudevan2020semantic} by using their best configuration. We select scenes $148$ of the test set. %
From Fig.~\ref{fig:qual} we clearly see that our method produces higher quality images from audio.

\def \qualfigwidth {0.14}
\begin{figure}[t]
    \begin{center}
    \subfigure[GT]{
        \includegraphics[width=\qualfigwidth\linewidth]{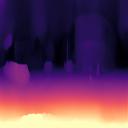}
    }
    \subfigure[\cite{vasudevan2020semantic}]{
        \includegraphics[width=\qualfigwidth\linewidth]{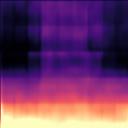}
    }
    \subfigure[Proposed]{
        \includegraphics[width=\qualfigwidth\linewidth]{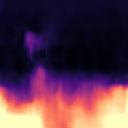}
    }
    \subfigure[GT]{
        \includegraphics[width=\qualfigwidth\linewidth]{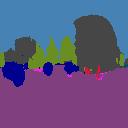}
    }
    \subfigure[\cite{vasudevan2020semantic}]{
        \includegraphics[width=\qualfigwidth\linewidth]{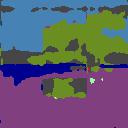}
    }
    \subfigure[Proposed]{
        \includegraphics[width=\qualfigwidth\linewidth]{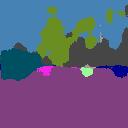}
    }\\
    \caption{Qualitative results for scenes $148$.} %
    \end{center}
    \label{fig:qual}
\end{figure}

By looking at the depth estimation results we observe that method \cite{vasudevan2020semantic} predict a similar depth pattern for all of the three tested audio segments. Although we small variation in scale, their method is not able to detect important details such objects at medium distance. Our method instead can identify those object in a more reliable way. Moreover, their depth prediction seems not capable at localizing objects because it mainly focuses at the central part of the image no matter what the ground truth is.

If we focus on the estimation of semantic segmentation maps, we observe that our method is by far superior compared to \cite{vasudevan2020semantic}. 
In fact, the method of \cite{vasudevan2020semantic} is affected by severe issues. First of all, their method miss classifies objects which should be easily detected due to the dataset statistics. For instance, if we consider the ``road'' class we see that their method confuses the ``road'' with ``tree'' for large regions, Fig.~\ref{fig:qual}(e). 
Another important issue is the ``unrealism'' of segmentation for certain classes. For example, if we consider the ``building'' class we see that \cite{vasudevan2020semantic} often predicts them both: in  shapes uncorrelated with respect to the ground truth data, and in positions not physically possible. 
Also the detection of moving objects such as ``cars'' is often undefined, and unrealistic as well, Fig.~\ref{fig:qual}(e).

In contrast, our method is able to: reproduce the main structure of the image, predict objects in appropriate locations, and maintain correspondence of classes between prediction and ground-truth. We provide more qualitative examples in the Supplementary Appendix.

\section{Conclusions}

We have presented a novel framework for estimating visual information from audio which outperforms the current state-of-the-art for estimating depth maps and semantic segmentation maps from multi channel audio.
The fundamental idea behind our proposed method is learning the transformation between the audio and the visual domains at the visual manifold level obtained using a VQ-VAE rather than using an end-to-end training approach. We have shown that this transformation is learnable and effective, and results in superior performance compared to previous approaches to this problem. Thanks to the quantized nature of the VQ-VAE's manifold, our method is more robust to errors and is able to generate more realistic images from audio only.

As future work, we plan to define a new performance metric that takes into account important visual properties which are not considered by traditional metrics for depth map estimation and semantic segmentation estimation. Based on this new metric, alternative optimization objectives can be investigated.
We also plan to collect a new indoor audio/visual dataset organized and recoreded in way which is more appropriate for estimating visual information from sound.

\bibliographystyle{splncs04}
\bibliography{macros, main}

\clearpage
\appendix

\setcounter{page}{1}

\begin{center}
\Large
\textbf{Estimating Visual Information From Audio
Through Manifold Learning} \\
\vspace{0.5em} Supplementary Material \\
\vspace{1.0em}
\end{center}
\appendix

\section{Qualitative examples}

We provide additional qualitative examples comparing between \cite{vasudevan2020semantic} and our method.
Our method provides improved depth and semantic segmentation from audio compared to \cite{vasudevan2020semantic}.

\def \qualfigwidth {0.14}
\begin{figure}
    \begin{center}
    \def\s{.4}
    \subfigure[GT]{\includegraphics[scale=\s]{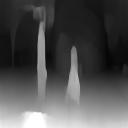}}
    \subfigure[\cite{vasudevan2020semantic}]{\includegraphics[scale=\s]{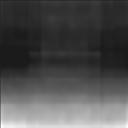}}
    \subfigure[Ours]{\includegraphics[scale=\s]{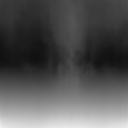}}
    \subfigure[GT]{\includegraphics[scale=\s]{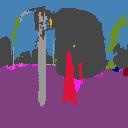}}
    \subfigure[\cite{vasudevan2020semantic}]{\includegraphics[scale=\s]{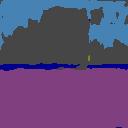}}
    \subfigure[Ours]{\includegraphics[scale=\s]{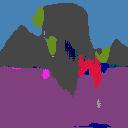}}
    \end{center}
    \caption{scene 140.}
    \label{fig:supp-140}
\end{figure}

\def \qualfigwidth {0.14}
\begin{figure}
    \begin{center}
    \def\s{.4}
    \subfigure[GT]{\includegraphics[scale=\s]{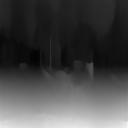}}
    \subfigure[\cite{vasudevan2020semantic}]{\includegraphics[scale=\s]{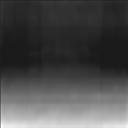}}
    \subfigure[Ours]{\includegraphics[scale=\s]{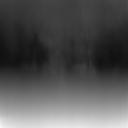}}
    \subfigure[GT]{\includegraphics[scale=\s]{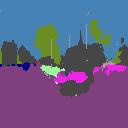}}
    \subfigure[\cite{vasudevan2020semantic}]{\includegraphics[scale=\s]{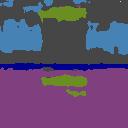}}
    \subfigure[Ours]{\includegraphics[scale=\s]{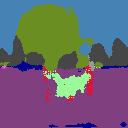}}
    \end{center}
    \caption{scene 141.}
    \label{fig:supp-141}
\end{figure}

\def \qualfigwidth {0.14}
\begin{figure}
    \begin{center}
    \def\s{.4}
    \subfigure[GT]{\includegraphics[scale=\s]{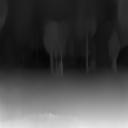}}
    \subfigure[\cite{vasudevan2020semantic}]{\includegraphics[scale=\s]{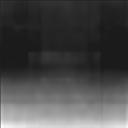}}
    \subfigure[Ours]{\includegraphics[scale=\s]{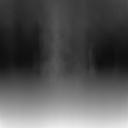}}
    \subfigure[GT]{\includegraphics[scale=\s]{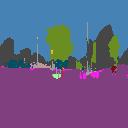}}
    \subfigure[\cite{vasudevan2020semantic}]{\includegraphics[scale=\s]{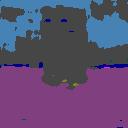}}
    \subfigure[Ours]{\includegraphics[scale=\s]{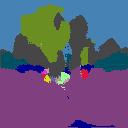}}
    \end{center}
    \caption{scene 142.}
    \label{fig:supp-142}
\end{figure}

\def \qualfigwidth {0.14}
\begin{figure}
    \begin{center}
    \def\s{.4}
    \subfigure[GT]{\includegraphics[scale=\s]{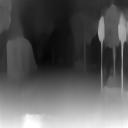}}
    \subfigure[\cite{vasudevan2020semantic}]{\includegraphics[scale=\s]{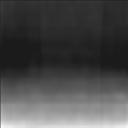}}
    \subfigure[Ours]{\includegraphics[scale=\s]{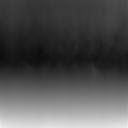}}
    \subfigure[GT]{\includegraphics[scale=\s]{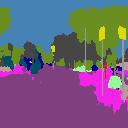}}
    \subfigure[\cite{vasudevan2020semantic}]{\includegraphics[scale=\s]{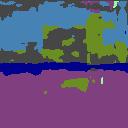}}
    \subfigure[Ours]{\includegraphics[scale=\s]{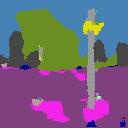}}
    \end{center}
    \caption{scene 143.}
    \label{fig:supp-143}
\end{figure}

\def \qualfigwidth {0.14}
\begin{figure}
    \begin{center}
    \def\s{.4}
    \subfigure[GT]{\includegraphics[scale=\s]{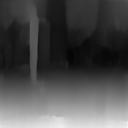}}
    \subfigure[\cite{vasudevan2020semantic}]{\includegraphics[scale=\s]{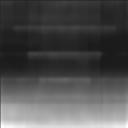}}
    \subfigure[Ours]{\includegraphics[scale=\s]{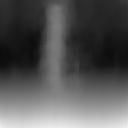}}
    \subfigure[GT]{\includegraphics[scale=\s]{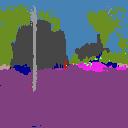}}
    \subfigure[\cite{vasudevan2020semantic}]{\includegraphics[scale=\s]{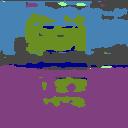}}
    \subfigure[Ours]{\includegraphics[scale=\s]{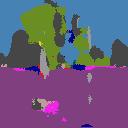}}
    \end{center}
    \caption{scene 144.}
    \label{fig:supp-144}
\end{figure}

\def \qualfigwidth {0.14}
\begin{figure}
    \begin{center}
    \def\s{.4}
    \subfigure[GT]{\includegraphics[scale=\s]{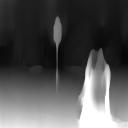}}
    \subfigure[\cite{vasudevan2020semantic}]{\includegraphics[scale=\s]{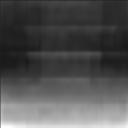}}
    \subfigure[Ours]{\includegraphics[scale=\s]{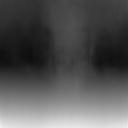}}
    \subfigure[GT]{\includegraphics[scale=\s]{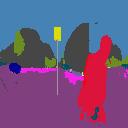}}
    \subfigure[\cite{vasudevan2020semantic}]{\includegraphics[scale=\s]{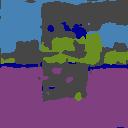}}
    \subfigure[Ours]{\includegraphics[scale=\s]{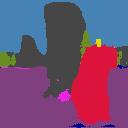}}
    \end{center}
    \caption{scene 145.}
    \label{fig:supp-145}
\end{figure}

\def \qualfigwidth {0.14}
\begin{figure}
    \begin{center}
    \def\s{.4}
    \subfigure[GT]{\includegraphics[scale=\s]{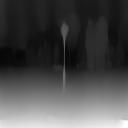}}
    \subfigure[\cite{vasudevan2020semantic}]{\includegraphics[scale=\s]{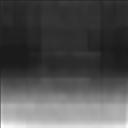}}
    \subfigure[Ours]{\includegraphics[scale=\s]{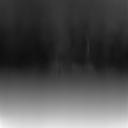}}
    \subfigure[GT]{\includegraphics[scale=\s]{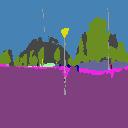}}
    \subfigure[\cite{vasudevan2020semantic}]{\includegraphics[scale=\s]{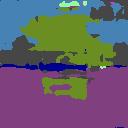}}
    \subfigure[Ours]{\includegraphics[scale=\s]{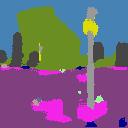}}
    \end{center}
    \caption{scene 146.}
    \label{fig:supp-146}
\end{figure}

\def \qualfigwidth {0.14}
\begin{figure}
    \begin{center}
    \def\s{.4}
    \subfigure[GT]{\includegraphics[scale=\s]{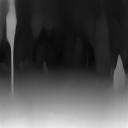}}
    \subfigure[\cite{vasudevan2020semantic}]{\includegraphics[scale=\s]{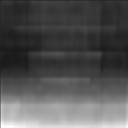}}
    \subfigure[Ours]{\includegraphics[scale=\s]{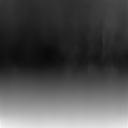}}
    \subfigure[GT]{\includegraphics[scale=\s]{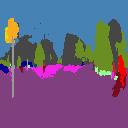}}
    \subfigure[\cite{vasudevan2020semantic}]{\includegraphics[scale=\s]{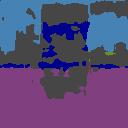}}
    \subfigure[Ours]{\includegraphics[scale=\s]{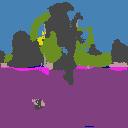}}
    \end{center}
    \caption{scene 147.}
    \label{fig:supp-147}
\end{figure}

\def \qualfigwidth {0.14}
\begin{figure}
    \begin{center}
    \def\s{.4}
    \subfigure[GT]{\includegraphics[scale=\s]{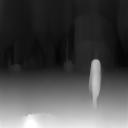}}
    \subfigure[\cite{vasudevan2020semantic}]{\includegraphics[scale=\s]{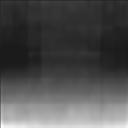}}
    \subfigure[Ours]{\includegraphics[scale=\s]{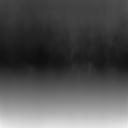}}
    \subfigure[GT]{\includegraphics[scale=\s]{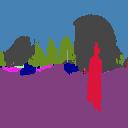}}
    \subfigure[\cite{vasudevan2020semantic}]{\includegraphics[scale=\s]{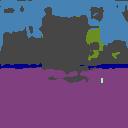}}
    \subfigure[Ours]{\includegraphics[scale=\s]{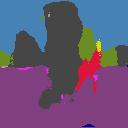}}
    \end{center}
    \caption{scene 148.}
    \label{fig:supp-148}
\end{figure}

\def \qualfigwidth {0.14}
\begin{figure}
    \begin{center}
    \def\s{.4}
    \subfigure[GT]{\includegraphics[scale=\s]{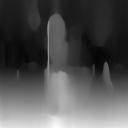}}
    \subfigure[\cite{vasudevan2020semantic}]{\includegraphics[scale=\s]{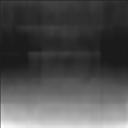}}
    \subfigure[Ours]{\includegraphics[scale=\s]{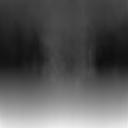}}
    \subfigure[GT]{\includegraphics[scale=\s]{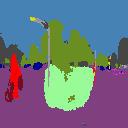}}
    \subfigure[\cite{vasudevan2020semantic}]{\includegraphics[scale=\s]{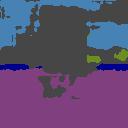}}
    \subfigure[Ours]{\includegraphics[scale=\s]{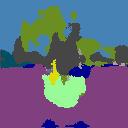}}
    \end{center}
    \caption{scene 149.}
    \label{fig:supp-149}
\end{figure}

\def \qualfigwidth {0.14}
\begin{figure}
    \begin{center}
    \def\s{.4}
    \subfigure[GT]{\includegraphics[scale=\s]{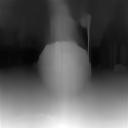}}
    \subfigure[\cite{vasudevan2020semantic}]{\includegraphics[scale=\s]{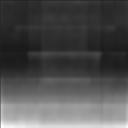}}
    \subfigure[Ours]{\includegraphics[scale=\s]{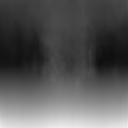}}
    \subfigure[GT]{\includegraphics[scale=\s]{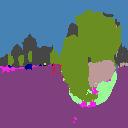}}
    \subfigure[\cite{vasudevan2020semantic}]{\includegraphics[scale=\s]{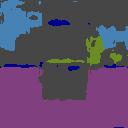}}
    \subfigure[Ours]{\includegraphics[scale=\s]{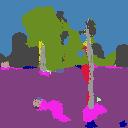}}
    \end{center}
    \caption{scene 150.}
    \label{fig:supp-150}
\end{figure}

\def \qualfigwidth {0.14}
\begin{figure}
    \begin{center}
    \def\s{.4}
    \subfigure[GT]{\includegraphics[scale=\s]{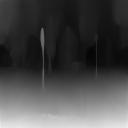}}
    \subfigure[\cite{vasudevan2020semantic}]{\includegraphics[scale=\s]{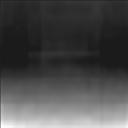}}
    \subfigure[Ours]{\includegraphics[scale=\s]{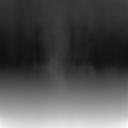}}
    \subfigure[GT]{\includegraphics[scale=\s]{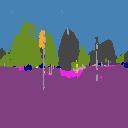}}
    \subfigure[\cite{vasudevan2020semantic}]{\includegraphics[scale=\s]{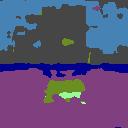}}
    \subfigure[Ours]{\includegraphics[scale=\s]{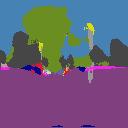}}
    \end{center}
    \caption{scene 151.}
    \label{fig:supp-151}
\end{figure}
\clearpage
\section{Failure examples}

\begin{figure}
    \begin{center}
    \def\s{.4}
    \subfigure[GT]{\includegraphics[scale=\s]{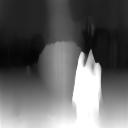}}
    \subfigure[Ours]{\includegraphics[scale=\s]{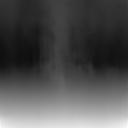}}
    \hspace{2mm}
    \subfigure[GT]{\includegraphics[scale=\s]{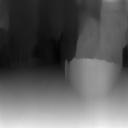}}
    \subfigure[Ours]{\includegraphics[scale=\s]{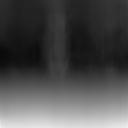}}
    \hspace{2mm}
    \subfigure[GT]{\includegraphics[scale=\s]{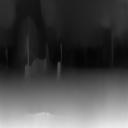}}
    \subfigure[Ours]{\includegraphics[scale=\s]{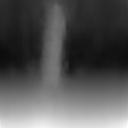}}
    \end{center}
    \caption{Depth estimation failure cases.}
    \label{fig:supp-fail-depth}
\end{figure}

\begin{figure}
    \begin{center}
    \def\s{.4}
    \subfigure[GT]{\includegraphics[scale=\s]{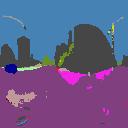}}
    \subfigure[Ours]{\includegraphics[scale=\s]{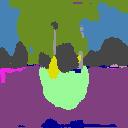}}
    \hspace{2mm}
    \subfigure[GT]{\includegraphics[scale=\s]{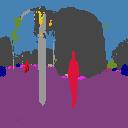}}
    \subfigure[Ours]{\includegraphics[scale=\s]{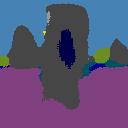}}
    \hspace{2mm}
    \subfigure[GT]{\includegraphics[scale=\s]{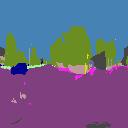}}
    \subfigure[Ours]{\includegraphics[scale=\s]{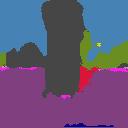}}
    \end{center}
    \caption{Semantic segmentation estimation failure cases.}
    \label{fig:supp-fail-seg}
\end{figure}

We provide some failure cases of the proposed method where: objects close to camera are not identified, Fig.~\ref{fig:supp-fail-depth}, objects are miss classified, Fig.~\ref{fig:supp-fail-seg}.

\def \qualfigwidth {0.14}

\def \qualfigwidth {0.14}

\end{document}